\documentclass[11pt,a4paper]{article}
\usepackage[hyperref]{naaclhlt2019}
\usepackage[numbers]{natbib}
\usepackage{booktabs}
\usepackage{times}
\usepackage{latexsym}
\usepackage{amsmath}
\usepackage{amsfonts}
\usepackage{mathtools}
\usepackage{url}
\usepackage{graphicx}
\aclfinalcopy 

\title{Dialogue Natural Language Inference}

\author{Sean Welleck \\
  New York University \\
  {\tt wellecks@nyu.edu} \\\And
  Jason Weston \\
  Facebook AI Research \\
  New York University \\\And
  Arthur Szlam \\
  Facebook AI Research \\\And
  Kyunghyun Cho \\
  New York University \\
  Facebook AI Research \\
  CIFAR Azrieli Global Scholar 
  \\}

\date{}

\begin{document}
\maketitle
\begin{abstract}
  
Consistency is a long standing issue faced by dialogue models. In this paper, we frame the consistency of dialogue agents as natural language inference (NLI) and create a new natural language inference dataset called Dialogue NLI. We propose a method which demonstrates that a model trained on Dialogue NLI can be used to improve the consistency of a dialogue model, and evaluate the method with human evaluation and with automatic metrics on a suite of evaluation sets designed to measure a dialogue model's consistency.
\end{abstract}

\section{Introduction}

A long standing issue faced by dialogue models is \textit{consistency} \cite{Li2016AModel, Vinyals2015AModel, Zhang2018PersonalizingToo}. An example from \cite{Vinyals2015AModel} shows a two-round dialogue in which their neural sequence model first responds to \textit{what is your job?} with \textit{i'm a lawyer}, then responds to \textit{what do you do?} with \textit{i'm a doctor}.
Even when inconsistencies are relatively rare and semantically plausible, they are jarring, and because semantic plausibility is not enough to root them out, preventing them is challenging.  

\begin{figure*}
\centering
\begin{minipage}{.49\textwidth}
  \includegraphics[width=0.95\linewidth]{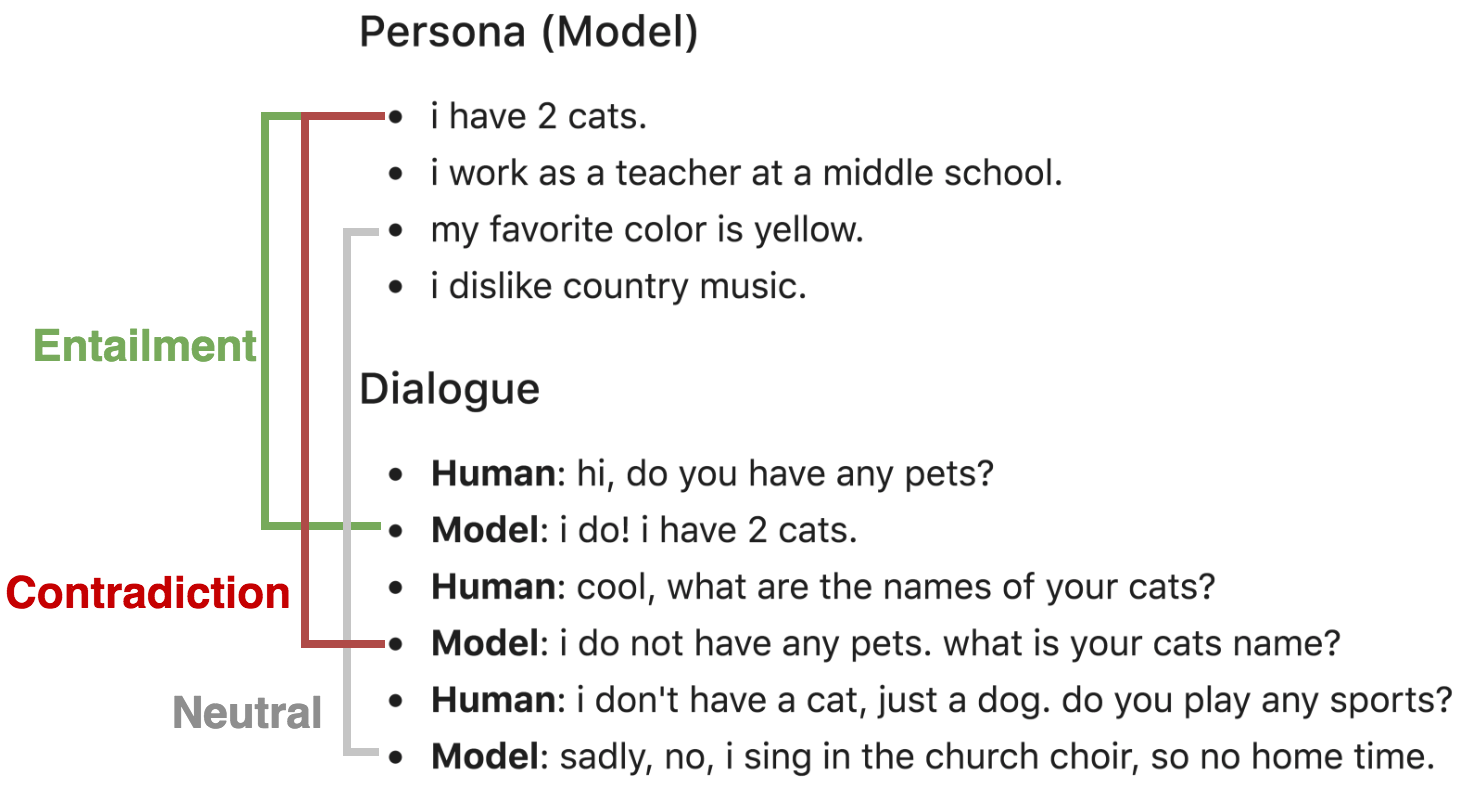}
  \captionof{figure}{Persona-based dialogue with a Key-Value Memory Network trained on Persona-Chat \cite{Zhang2018PersonalizingToo}.}
  \label{fig:chat}
\end{minipage}%
\hfill
\begin{minipage}{.49\textwidth}
  \includegraphics[width=0.95\linewidth]{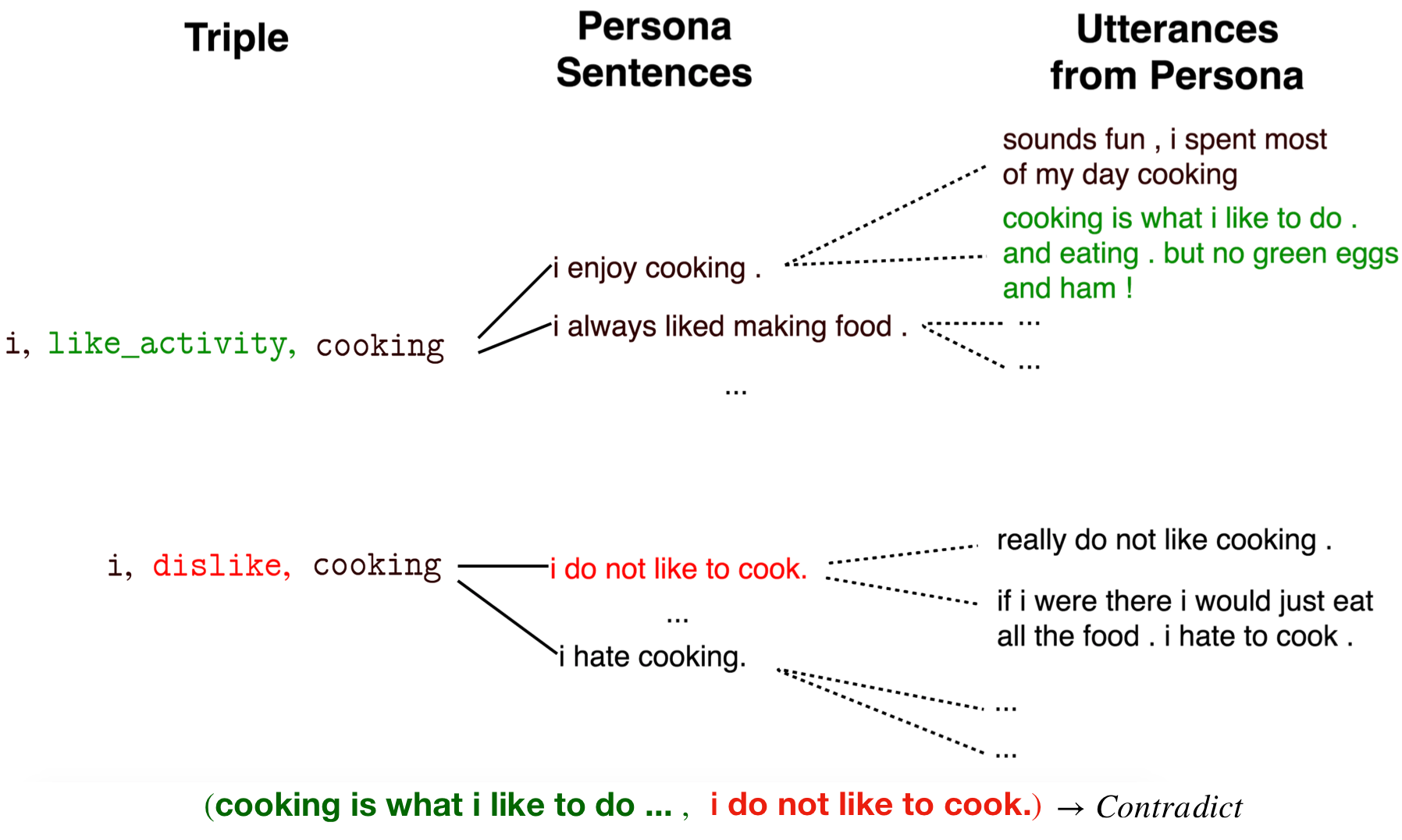}
  \captionof{figure}{Relating triples, persona sentences, and utterances to derive annotated sentence pairs. Shown here is a ``relation swap'' contradiction.}
  \label{fig:annot}
\end{minipage}%
\end{figure*}

One approach to increasing the consistency of a chit-chat dialogue model was proposed in \cite{Zhang2018PersonalizingToo}, where the dialogue agent was given a set of personal facts describing its character (a {\it persona}) and produces utterances that reflect the persona.  The intended outcome is that the agent produces utterances consistent with its given persona.
However, these models still face the consistency issue, as shown in Figure \ref{fig:chat}. 

Separately, the framework of Natural Language Inference (NLI) \cite{Bowman2015AInference,Dagan2006TheChallenge,Maccartney2009AnLogic} involves learning a mapping between a sentence pair and an entailment category. It is hypothesized that the NLI task is a proxy for general goals in natural language processing, such as language understanding \cite{Bowman2015AInference,Williams2018AInference}. Thus, the NLI task has been used for learning general sentence representations \cite{Conneau2017SupervisedData} and for evaluating NLP models \cite{Poliak2018OnInference,Wang2018GLUE:Understanding}, with the expectation that such models will be useful in downstream tasks.

Despite this expectation, leveraging an NLI model for a downstream task remains  an under-explored research direction. An NLI model may improve downstream task performance if properly used, while downstream tasks may yield new datasets or identify issues with existing NLI models, thus expanding the NLI research domain.

In this paper, we reduce the problem of consistency in dialogue to natural language inference. We first create a dataset, Dialogue NLI,\footnote{The dataset is available at \url{wellecks.github.io/dialogue_nli}.}
which contains sentence pairs labeled as entailment, neutral, or contradiction. 

Then, we demonstrate that NLI can be used to improve the consistency of dialogue models using a simple method where utterances are re-ranked using a NLI model trained on Dialogue NLI. The method results in fewer persona contradictions on three evaluation sets. The evaluation sets can be used independently to automatically evaluate a dialogue model's persona consistency, reducing the need for human evaluation. We discuss several future research directions involving this approach. 

\section{Dialogue Consistency and Natural Language Inference} 

First, we review the dialogue generation and natural language inference problems as well as the notions of consistency used throughout. 

\paragraph{Dialogue Generation} Dialogue generation can be framed as \textit{next utterance prediction}, in which an utterance (a sequence of tokens representing a sentence) $u_{t+1}$ is predicted given a conversation prefix $u_{\leq t}$. 
A sequence of utterances is interpreted as a \textit{dialogue} between \textit{agents}. For instance, an alternating two-agent dialogue which starts with agent $A$ and ends with agent $B$ is written as $u_{1}^{A}, u_{2}^{B}, u_{3}^{A}, u_{4}^{B},...,u_{T}^{B}$.

\paragraph{Persona-Based Dialogue} In \textit{persona-based dialogue}, each agent is associated with a persona, $P_{A}$ and $P_{B}$. An utterance is now predicted using the conversation prefix $u_{\leq t}$ \textit{and} the agents own persona, e.g. $P_{A}$ for agent $A$. It is assumed that an agent's utterances are conditionally dependent on its persona, which can be interpreted as the utterances being representative of, or reflecting, the persona. 

A typical approach for representing the persona is to use a set of sentences $P=\lbrace{p_1,...,p_m\rbrace}$.

\paragraph{Consistency} 

A \textit{consistency error}, or contradiction, occurs when an agent produces an utterance that contradicts one of their previous utterances. Similarly, a \textit{persona consistency error}, or persona contradiction, occurs when an agent produces an utterance that contradicts a subset of its persona.

A contradiction may be a clear logical contradiction, e.g. \textit{I have a dog} vs. \textit{I do not have a dog}, but in general is less clearly defined. 
As a result, in addition to logical contradictions, 
we interpret a consistency error as being two utterances not likely to be said by the same persona. For instance, ``i'm looking forward to going to the basketball game this weekend!'' vs. ``i don't like attending sporting events'', as well as ``i'm a lawyer'' vs. ``i'm a doctor'' would be viewed here as contradictions, although they are not strict logical inconsistencies. 

Similarly, a persona consistency error is interpreted here as an utterance which is not likely to be said given a persona described by a given set of persona sentences, in addition to logical contradictions. 

\paragraph{Natural Language Inference} Natural Language Inference (NLI) assumes a dataset $\mathcal{D}=\lbrace{(s_1,s_2)_i, y_i \rbrace}_{i=1}^{N}$ which associates an input pair $(s_1,s_2)$ to one of three classes $y\in \lbrace{\text{entailment}, \text{neutral}, \text{contradiction}\rbrace}$. Each input item $s_j$ comes from an input space $\mathcal{S}_{j}$, which in typical NLI tasks is the space of natural language sentences, i.e. $s_j$ is a sequence of words $(w_1,...,w_K)$ where each word $w_k$ is from a vocabulary $\mathcal{V}$. 

The input $(s_1,s_2)$ are referred to as the \textit{premise} and \textit{hypothesis}, respectively, and each label is interpreted as meaning the premise \textit{entails} the hypothesis, the premise is \textit{neutral} with respect to the hypothesis, or the premise \textit{contradicts} the hypothesis. The problem is to learn a function $f_{\text{NLI}}(s_1,s_2)\rightarrow \lbrace{E,N,C\rbrace}$ which generalizes to new input pairs.

\paragraph{Reducing Dialogue Consistency to NLI} 

Identifying utterances which contradict previous utterances or an agent's persona can be reduced to natural language inference by assuming that contradictions are contained in a sentence pair. 
That is, given a persona $P_A=\lbrace{p_1^A,...,p_m^A\rbrace}$ for agent A and a length-$T$ dialogue $u_1^A,u_2^B,...u_{T-1}^A,u_T^B$, it is assumed that a dialogue contradiction for agent A is contained in an utterance pair $(u_i^A, u_j^A)$, and a persona contradiction is contained in a pair $(u_i^A, p_k^A)$. Similarly, we assume that entailments and neutral interactions, defined in Section \ref{sec:dataset}, are contained in sentence pairs. We do not consider relationships which require more than two sentences to express.

Under this assumption, we can use a natural language inference model $f_{\text{NLI}}$ to identify entailing, neutral, or contradicting utterances. 

Section~\ref{sec:dataset} proposes a dialogue-derived dataset for training $f_{\text{NLI}}$, and Section \ref{sec:rerank} proposes a method which incorporates $f_{\text{NLI}}$ with a dialogue model for next utterance prediction. 

\section{Dialogue NLI Dataset}
\label{sec:dataset}

The Dialogue NLI dataset consists of sentence pairs labeled as entailment (E), neutral (N), or contradiction (C). 

\paragraph{Sentences} 

Sentences originate from a two-agent persona-based dialogue dataset. 
A dialogue between agents $A$ and $B$ consists of a sequence of utterances $u_{1}^{A}, u_{2}^{B}, u_{3}^{A}, u_{4}^{B},...,u_{T}^{B}$, and each agent has a persona represented by a set of persona sentences $\lbrace{p_1^A,...,p_{m_A}^A\rbrace}$ and $\lbrace{p_1^B,...,p_{m_B}^B\rbrace}$. The Dialogue NLI dataset consists of $(u_i, p_j)$ and $(p_i,p_j)$ pairs\footnote{
We also release additional $(u_i,u_j)$ pairs, but experiments in this paper are not based on them.
} from the Persona-Chat dataset~\cite{Zhang2018PersonalizingToo}\footnote{The dataset collection process is applicable to other persona-based dialogue datasets such as \cite{Mazare2018TrainingAgents}.}. 

\subsection{Triple Generation} 

In order to determine labels for our dataset, we require human annotation of the utterances and persona sentences in PersonaChat, as the original dataset does not contain this information. We perform such annotation by first associating a human-labeled \textit{triple} $(e_1,r,e_2)$ with each persona sentence, and a subset of all the utterances, detailed in \ref{ssec:triple-annotate}. Each triple contains the main fact conveyed by a persona sentence, such as $\texttt{(i, have\_pet, dog)}$ for the persona sentence \textit{I have a pet dog}, or a fact mentioned in an utterance, such as \textit{No, but my dog sometimes does}. 

Persona sentences and utterances are grouped by their triple (e.g. see Figure \ref{fig:annot}),
and pairs $(u,p)$ and $(p,p)$ are defined as entailment, neutral, or contradiction based on their triple according to the criteria below. For examples and summary, we refer readers to Tables~\ref{tbl:examples}--\ref{tbl:pnliprop}.

\begin{table*}[t!]
\small
\begin{center}
\resizebox{\linewidth}{!}{
\begin{tabular}{lp{3cm}|p{3cm}r|c}
\bf Triple &   \bf Premise & \bf Hypothesis  & \bf Triple  & \bf Label  \\ 
\toprule
(i, like\_activity, chess) & i listen to a bit of everything . it helps me focus for my chess tournaments . & i like to play chess . & (i, like\_activity, chess) & E \\
\midrule
- &	how are you today? & i drink espresso . & (i, like\_drink, espresso) & N \\
\midrule
(i, like\_goto, spain) & i love spain so much , i been there 6 times . & i think i will retire in a few years . & (i, want\_do, retire) & N \\
\midrule
(i, have\_vehicle, car) & my vehicle is older model car . & i have pets .  & (i, have\_pet, pets) & N \\
\midrule
(i, dislike, cooking) &	i really do not enjoy preparing food for myself . & i like to cook with food i grow in my garden . & (i, like\_activity, cooking) & C \\
\midrule
(i, physical\_attribute, short)	& height is missing from my stature . &	i am 7 foot tall . &	(i, physical\_attribute, tall) & C \\
\midrule
(i, have\_family, 3 sister)  & i have a brother and 3 sisters . & i have a brother and four sisters . & (i, have\_family, 4 sister) & C \\
\end{tabular}
}
\end{center}
\caption{\label{tbl:examples} Examples from the validation set.}
\end{table*}

\begin{table*}[t!]
\begin{center}
\begin{tabular}{lr|cc|cc|cc|cc}
  & & \multicolumn{2}{c|}{\bf Train} &  \multicolumn{2}{c|}{\bf Valid} & \multicolumn{2}{c}{\bf Test} & \multicolumn{2}{c}{\bf Test-Gold}\\

\textbf{Data Type} & \textbf{Label} & $(u,p)$ & $(p,p)$ & $(u,p)$ & $(p,p)$ & $(u,p)$ & $(p,p)$ & $(u,p)$ & $(p,p)$ \\
\toprule
Matching Triple & E & 43,000 & 57,000 & 5,000 & 500 & 4,500 & 900 & 3,712 & 615 \\ 
\midrule
Misc. Utterance & N & 50,000 & -      & 3,350 & -   & 3,000 & - & 2,282 & -  \\ 
Persona Pairing & N & 20,000 & 10,000 & 2,000 & -   & 2,000 & - & 1,466 & -  \\ 
Relation Swap   & N & 20,000 & -      & 150   & -   & 400   & - & 260 & -  \\ 
\midrule
Relation Swap  & C  & 19,116 & 2,600  & 85    & 14  & 422   & 50  & 279 & 44 \\ 
Entity Swap    & C  & 47,194 & 31,200 & 4,069 & 832 & 3,400 & 828 & 2,246 & 591 \\ 
Numerics       & C  & 10,000 & - & 500 & - & 1,000 & - & 881 & - \\ 
\midrule
\multicolumn{2}{c|}{\textbf{Dialogue NLI Overall}} & \multicolumn{2}{c|}{\textbf{310,110}} & \multicolumn{2}{c|}{\textbf{16,500}} & \multicolumn{2}{c|}{\textbf{16,500}} &
\multicolumn{2}{c}{\textbf{12,376}}\\
\end{tabular}
\end{center}
\caption{\label{tbl:pnliprop} Dialogue NLI Dataset Properties. $(u,p)$ and $(p,p)$ refer to (utterance, persona sentence) and (persona sentence, persona sentence) pairs, respectively. Numerics consist of $(u,u)$ $(u,p)$ and $(p,p)$ pairs. }
\end{table*}

\paragraph{Entailment} 
Each unique pair of sentences that share the same triple are labeled as entailment. 

\paragraph{Neutral} 

Neutral pairs are obtained with three different methods. 

First, a {\em{miscellaneous utterance}} is a $(u, p)$ pair of which $u$ is not associated with any triple. This includes greetings (\textit{how are you today?}) and sentences unrelated to a persona sentence (\textit{the weather is ok today}), so such utterances are assumed to be neutral with respect to persona sentences.

The second method, {\em{persona pairing}}, takes advantage of the fact that each ground-truth persona is typically neither redundant nor contradictory. A persona sentence pair $(p, p')$ is first selected from a persona if $p$ and $p'$ do not share the same triple. Then each sentence associated with the same triple as $p$ is paired with each sentence associated with the same triple as $p'$. 

Lastly, we specify {\em{relation swaps}} $(r, r')$ for certain relations (see Appendix~\ref{apx:relswap}) whose triples are assumed to represent independent facts, such as $\texttt{have\_vehicle}$ and $\texttt{have\_pet}$. A sentence pair, whose first sentence is associated with a triple $(\cdot, r, \cdot)$ and whose second sentence has triple $(\cdot, r', \cdot)$, is labeled as neutral. See Table \ref{tbl:examples} for an example. 

\paragraph{Contradiction}

We obtain contradictions using three methods. See Figure~\ref{fig:annot} for an example.

First, the {\em{relation swap}} method is used by specifying contradicting relation pairs $(r, r')$ (see Appendix \ref{apx:relswap}), such as $(\texttt{like\_activity}, \texttt{dislike})$, then pairing each sentence associated with the triple $(e_1,r,e_2)$ with each sentence associated with $(e_1,r',e_2)$. 

Similarly, an {\em{entity swap}} consists of specifying relations, e.g., $\texttt{physical\_attribute}$, that would yield a contradiction when the value of $e_2$ is changed to a different value $e_2'$, e.g., $\texttt{short}\rightarrow \texttt{tall}$ (see Appendix \ref{apx:eswap}). Sentences associated with $(e_1,r,e_2)$ are then paired with sentences associated with $(e_1,r,e_2')$.

Finally, a {\em{numeric}} contradiction is obtained by first selecting a sentence which contains a number that appears in the associated triple (see Table \ref{tbl:examples}). A contradicting sentence is generated by replacing the sentence's numeric surface form with a different randomly sampled integer in the number or text form.

\subsection{Triple Annotation}
\label{ssec:triple-annotate} 

Each persona sentence is annotated with a triple $(e_1,r,e_2)$ using Amazon Mechanical Turk task. We first define a schema consisting of $\langle category\rangle \langle relation\rangle \langle category\rangle$ rules, such as $\langle person\rangle have\_pet \langle animal\rangle$, where the relation comes from a fixed set of relation types $\mathcal{R}$, listed in Appendix \ref{apx:schema}. Given a sentence, the annotator selects a relation $r$ from a drop-down populated with the values in $\mathcal{R}$. The annotator then selects the categories and values of the entities $e_1$ and $e_2$ using drop-downs that are populated based on the schema rules. An optional drop-down contains numeric values for annotating entity quantities (e.g., 3 brothers). If selected, the numeric value is concatenated to the front of the entity value. The annotator can alternatively input an out-of-schema entity value in a text-box. 
Using this method, each of the 10,832 persona sentences is annotated with a triple $(e_1,r,e_2)$, where $r\in\mathcal{R}$, $e_1\in\mathcal{E}_1$, and $e_2\in\mathcal{E}_2$. Here $\mathcal{E}_1$ is the set of all annotated $e_1$ from the drop-downs or the text-box, and $\mathcal{E}_2$ is similarly defined.

Finally, \textit{utterances} are associated with a triple as follows. Let $p$ be a persona sentence with triple $(e_1,r,e_2)$. We start with all utterances, $U$, from agents that have $p$ in their persona. An utterance $u\in U$ is then associated with the triple $(e_1,r,e_2)$ and persona sentence $p$ when $e_2$ is a sub-string of $u$, or word similarity\footnote{
We use cosine similarity between the mean of TF-IDF weighted GloVe~\cite{Pennington2014GloVe:Representation} word vectors and set $\tau=0.9$.
}  $\text{sim}(u, p)\geq \tau$ is suitably large.

\subsection{Statistics}

Table~\ref{tbl:pnliprop} summarizes the dataset and its underlying data types. The label, triple, and data type are supplied as annotations for each sentence pair. We additionally create a gold-standard test set (\textit{Test Gold}) by crowdsourcing three label annotations for each example in the test set. We keep each test example for which two or more annotators agreed with its dataset label. All sentences in Dialogue NLI were generated by humans during the crowdsourced dialogue collection process of the Persona-Chat dataset \cite{Zhang2018PersonalizingToo}. The resulting sentence pairs are thus drawn from a natural dialogue domain that differs from existing NLI datasets, which are either drawn from different domains such as image captions or created using synthetic templates~\cite{Bowman2015AInference,Demszky2018TransformingDatasets,Khot2018SCITAIL:Answering,Marelli2014AModels, Poliak2018CollectingEvaluation, Wang2018GLUE:Understanding,Williams2018AInference}.

\section{Consistent Dialogue Agents via Natural Language Inference}
\label{sec:rerank} 

We now present a method which demonstrates that natural language inference can be used to improve 
the consistency of dialogue agents. Candidate utterances are re-ranked based on whether the candidate is predicted to contradict a persona sentence. If the NLI model predicts that a candidate contradicts a persona sentence, the candidate's score is penalized, with the penalty weighted by the NLI model's confidence\footnote{
In
our experiments, the softmax output corresponding to the contradiction class from Dialogue NLI.
} 
scaled by a constant. 

Specifically, assume a dialogue model $f^{\text{dialogue}}(P, u_{\leq t}, U)\rightarrow (s_1,s_2,...,s_{|U|})$ and a Dialogue NLI model $f^{\text{NLI}}(u,p)\rightarrow \lbrace{E, N, C\rbrace}$.
Given a persona  $P=\lbrace{p_1,...,p_m\rbrace}$, previous utterances $u_{\leq t}$, and a set of candidate next-utterances $U$, the dialogue model outputs a ranked list of scores $s_1,s_2,...,s_{|U|}$ corresponding to next-utterance candidates $u_1,u_2,...,u_{|U|}$. 

The NLI model is then run on each $(u_i, p_j)$ pair, predicting a label  $y_{i,j}\in \lbrace{E,N,C\rbrace}$ with confidence $c_{i,j}$. A contradiction score is computed for each candidate as:
\begin{equation*}
    s_i^\text{contradict} =
    \begin{cases*}
      0, & if $y_{i,j}\neq C\ \forall\ p_j\in P$ \\
      \max\limits_{j:y_{i,j}=C} c_{i,j},  & otherwise.
    \end{cases*}
\end{equation*}

That is, if the candidate $u_i$ does not contradict any persona sentence $p_j$ according to the NLI model, $s_i^\text{contradict}$ is zero. If $u_i$ contradicts one or more persona sentences, $s_i^\text{contradict}$ is the highest confidence, $c_{i,j}$, out of the contradicting $(u_i,p_j).$\footnote{
Future work could consider filtering previous-utterance contradictions $(u_i,u_j)$ as well.
}

New candidate scores are then computed as
\begin{equation}
    s_i^\text{re-rank} = s_i-\lambda(s_1 - s_k)s_i^{\text{contradict}}
\end{equation}
and the candidates are sorted according to $s^\text{re-rank}$. Hyper-parameters $\lambda$ and $k$ control the NLI model's influence in re-ranking. For example, if the top candidate has a contradiction score of $1.0$, then with $\lambda = 1$, it will be moved to the $k$'th position in the ranking. $\lambda = 0$ corresponds to no re-ranking.

\section{Experiments}
\label{expr:nli}

\subsection{Experiment 1: NLI}

\begin{table}[t!]
\begin{center}
\begin{tabular}{l|llp{5mm}}
\bf Model  & \bf Valid & \bf Test & \bf Test Gold \\ 
\toprule
ESIM                    & \textbf{86.31} & \textbf{88.20} & \textbf{92.45} \\
InferSent               & 85.82 & 85.68 & 89.96\\
\midrule
InferSent SNLI          & 47.86 & 46.36 & 47.03\\
InferSent Hyp. Only     & 55.98 & 57.19 & 51.52\\
Most Common Class       & 33.33 & 34.54 & 34.96\\
\midrule
ESIM Gold Triples           & 99.52 & 99.46 & 99.69\\ 
\end{tabular}
\end{center}
\caption{\label{tbl:pnli} Dialogue NLI Results }
\end{table}

\paragraph{Models} 

Many recently proposed NLI models can be categorized into sentence encoding based methods of the form $f_{\text{MLP}}(g_{\text{enc}}(s_1),g_{\text{enc}}(s_2))$, and attention-based methods of the form $f_{\text{MLP}}(g_{\text{attn}}(s_1, s_2))$ \cite{Lan2018NeuralAnswering}.
We thus choose and train representative models of each type which have achieved competitive performance on existing NLI benchmark datasets. For the sentence encoding method, we use InferSent~\cite{Conneau2017SupervisedData}, which encodes a sentence using a bidirectional LSTM followed by max-pooling over the output states. As the representative attention-based method we use the enhanced sequential inference model (ESIM, \cite{Chen2017EnhancedInference}), which computes an attention score for each word pair.

\label{expr:dialogue}
\begin{table*}[t!]
\begin{center}
\begin{tabular}{c|cc|cc|cc}
& \multicolumn{2}{c|}{Haves} & \multicolumn{2}{c}{Likes} & \multicolumn{2}{|c}{Attributes} \\
&   \bf Orig.  & \bf Rerank  & \bf Orig.  & \bf Rerank  & \bf Orig.  & \bf Rerank  \\
\toprule
Hits@1 $\uparrow$         & 30.2 & \textbf{37.3} & 16.9 & \textbf{18.7} & 35.2 & \textbf{36.4}  \\
Contradict@1 $\downarrow$ & 32.5 & \textbf{8.96} & 17.6 & \textbf{4.1} & 8.0 & \textbf{5.7}  \\
Entail@1 $\uparrow$       & 55.2 & \textbf{74.6} & 77.9 & \textbf{90.6} & 87.5 & \textbf{88.6} \\
\end{tabular}
\end{center}
\caption{\label{tbl:rerank} Effect of NLI re-ranking on persona consistency in dialogue. The reported metrics are percentages computed over each validation set.}
\end{table*}

We also report results from a model trained and evaluated using the hypothesis sentence only (InferSent Hyp. Only)~\cite{ Gururangan2018AnnotationData,Poliak2018HypothesisInference}, a model trained on the existing SNLI dataset \cite{Bowman2015AInference} but evaluated on Dialogue NLI (InferSent SNLI), and a model which returns the most common class from the Dialogue NLI training set (Most Common Class).

\paragraph{Results} 

Table~\ref{tbl:pnli} shows the performance of the two NLI models and three baselines on the Dialogue NLI validation and test sets. 
The test performance of ESIM (88.2\%) and InferSent (85.68\%) is similar to the performance reported on the existing SNLI dataset (88.0\% \cite{Chen2017EnhancedInference} and 85.5\% \cite{Conneau2017SupervisedData}, respectively), while the results on the Dialogue NLI gold test set (92.45\%, 89.96\%) are higher. As in Table \ref{tbl:pnli}, however, an InferSent model trained on SNLI performs poorly when evaluated on the proposed Dialogue NLI (47.03\%). This is likely due to a mismatch in sentence distributions between SNLI, which is derived from image captions, and Dialogue NLI, whose sentences more closely resemble downstream dialogue applications.
The hypothesis-only performance (51.52\%) is lower than the hypothesis-only baseline for SNLI (69.00\% \cite{Poliak2018HypothesisInference}), and shows that using information from both the utterance and persona sentence is necessary to achieve good performance on Dialogue NLI.

ESIM's reasonably strong performance on Dialogue NLI suggests that the model may be useful in a downstream task - a claim which we verify in Experiment~\ref{expr:dialogue}. However, there is also room for improvement. In particular, we report the performance of a model which takes the ground-truth triples as input instead of sentences. As shown in the last row of Table \ref{tbl:pnli}, each sentence's underlying triple contains sufficient information to achieve near-perfect accuracy (99.69\%). 

\begin{table*}[t!]
\begin{center}
\begin{tabular}{l|cc|cc|cc}
& \multicolumn{2}{c|}{Overall Score $\uparrow$} & \multicolumn{2}{c}{\% Consistent $\uparrow$} & \multicolumn{2}{|c}{\% Contradiction $\downarrow$} \\
&    Raw  & Calibrated  & Raw  & Calibrated  &  Raw  & Calibrated  \\
\toprule
KV-Mem              & 2.11$\pm$ 1.12 & 2.21$\pm$ 0.26 & 0.24 & 0.27$\pm$ 0.07 & 0.23 & 0.25$\pm$ 0.08 \\
KV-Mem + NLI        & \textbf{2.34$\pm$ 1.21} &\textbf{2.38$\pm$ 0.26} & \textbf{0.28} & \textbf{0.35$\pm$ 0.08} &\textbf{0.19} & \textbf{0.16$\pm$ 0.06} \\
\end{tabular}
\end{center}
\caption{\label{tbl:human} Human evaluation results (mean$\pm$ standard deviation).}
\end{table*}

\subsection{Experiment 2: Consistency in Dialogue}

This experiment evaluates the effect of the re-ranking method from Section~\ref{sec:rerank} on the dialogue model's persona consistency.

\paragraph{Experiment Setup} 

The re-ranking method of Section~\ref{sec:rerank} uses a dialogue next utterance prediction model
and the Dialogue NLI model.

For the dialogue model we train the key-value memory network~\cite{Zhang2018PersonalizingToo} on the Persona-Chat dataset, which uses persona sentences and the conversation prefix as context. This model achieved the best performance on Persona-Chat in \cite{Zhang2018PersonalizingToo}.
For the NLI model we use the ESIM model trained on Dialogue NLI, based on the results of Experiment \ref{expr:nli}.

To study the effect of re-ranking on persona consistency, we form  evaluation sets which contain next-utterances which are likely to yield persona contradiction or entailment, as follows. 

\paragraph{Evaluation Sets} 

\begin{figure*}
\begin{minipage}{.4\textwidth}
  \includegraphics[width=0.95\linewidth]{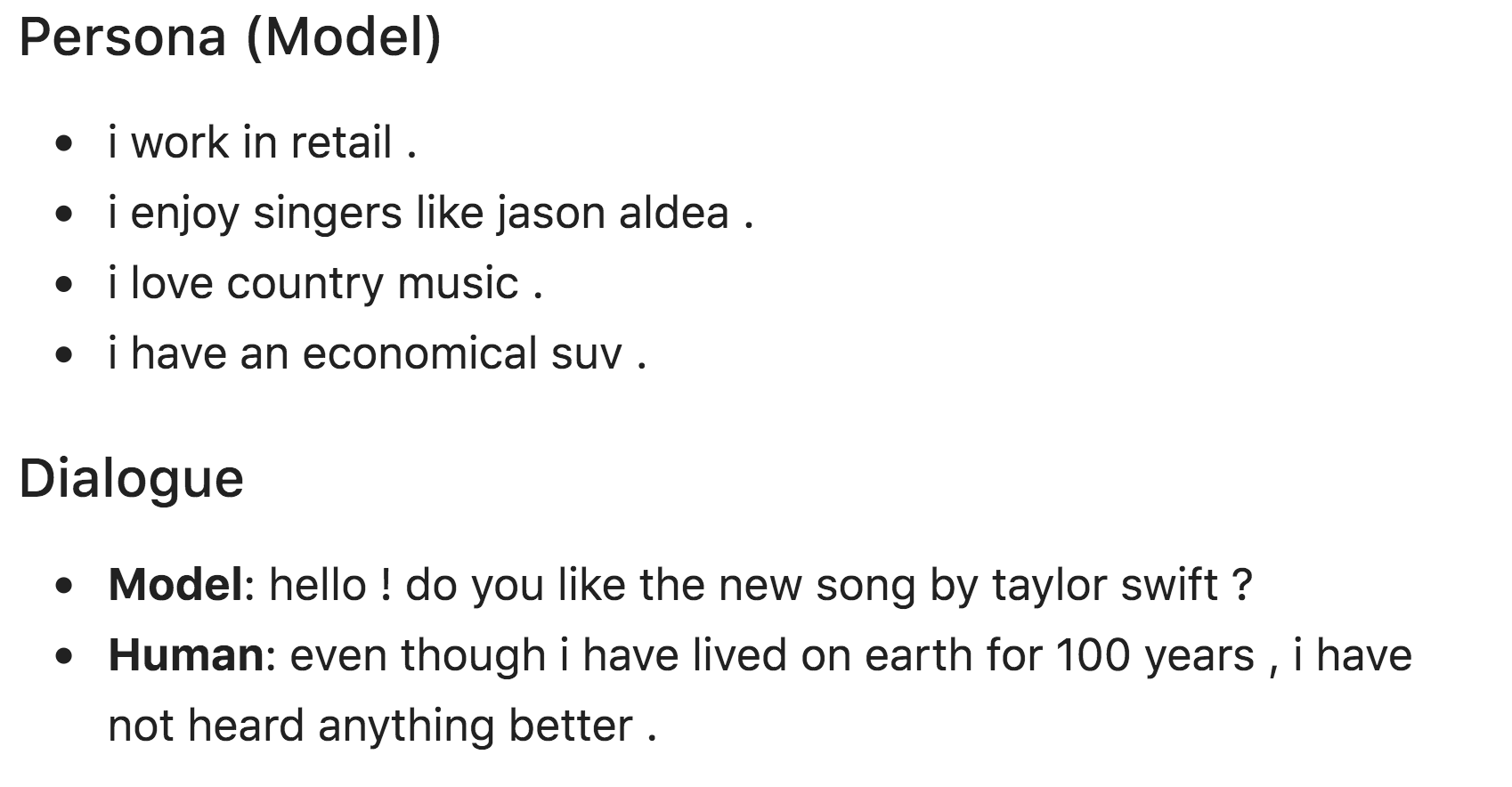}
\end{minipage}%
\begin{minipage}{.6\textwidth}
  \includegraphics[width=1.0\linewidth]{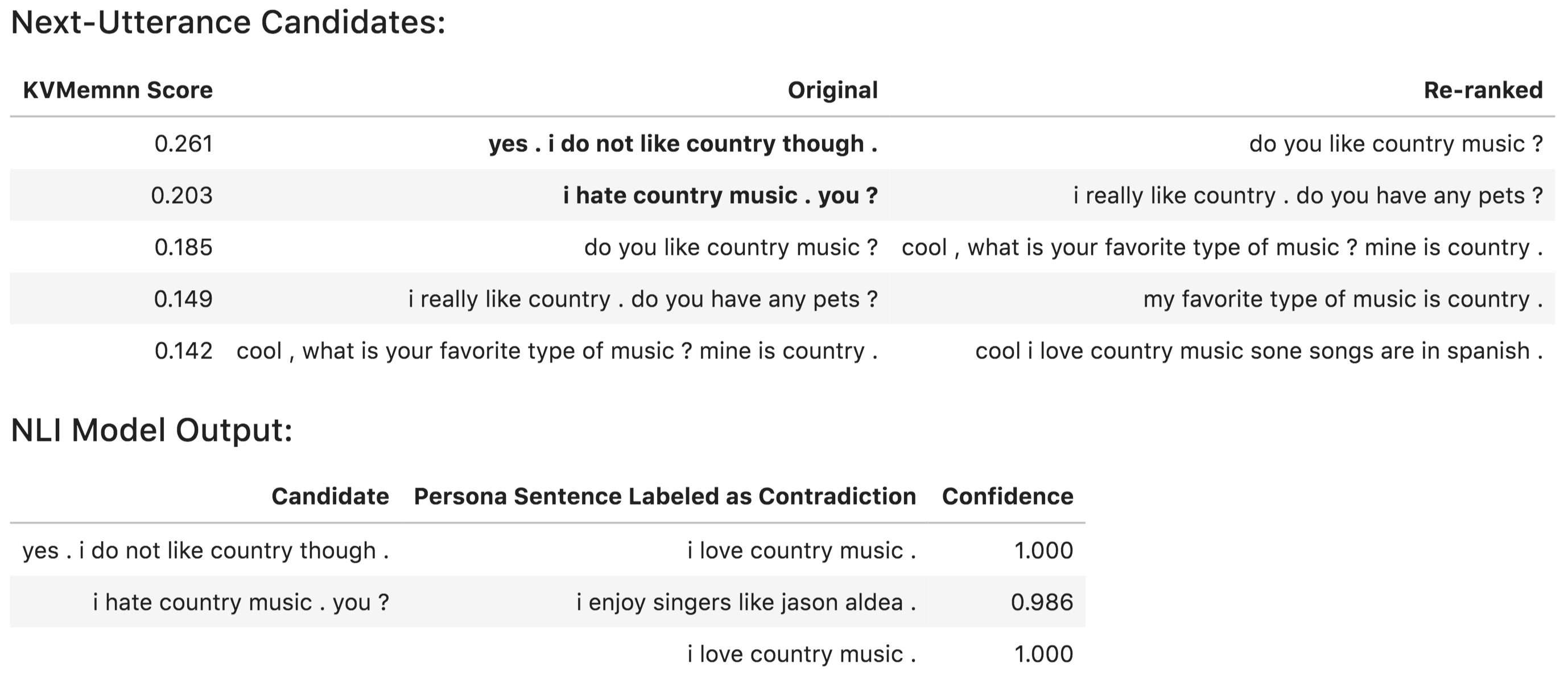}
\end{minipage}%
\caption{Example from the Likes Evaluation Set, showing dialogue model candidates, NLI model predictions, and reranked candidates using the method proposed in Section \ref{sec:rerank}.}
\label{fig:example}
\end{figure*}

Each example is formed by first finding a next-utterance $u_{t+1}$ in the Persona-Chat validation set which has an associated triple $(e_1,r,e_2)$ of interest, e.g. $(\texttt{i}, \texttt{like\_music}, \texttt{country})$. If a sentence in the agent's profile $P$ has triple $(e_1,r,e_2)$, we form the validation example $(P, u_{\leq t}, u_{t+1})$. Figure~\ref{fig:example} shows an example.

Each example is associated with candidates $U$, consisting of the ground-truth utterance $u_{t+1}$, 10 entailment candidates with the same triple as $u_{t+1}$, 10 contradicting candidates with a different triple than that of $u_{t+1}$, and 10 random candidates. The dialogue model must avoid ranking a contradicting candidate highly.

Specifically, suppose the ground-truth next-utterance $u_{t+1}$ is associated with triple $(e_1,r,e_2)$, e.g., $(\texttt{i}, \texttt{have\_pet}, \texttt{dog})$. Entailment candidates are utterances $u$ from the validation or training sets such that $u$ is associated with triple $(e_1,r,e_2)$. Since by construction a sentence in the profile also has triple $(e_1,r,e_2)$, these candidates entail a profile sentence. A contradicting candidate is an utterance associated with a specified contradicting triple $(e_1',r',e_2')$, e.g., $(\texttt{i}, \texttt{not\_have}, \texttt{dog})$.

We construct three evaluation sets, \textbf{Haves}, \textbf{Likes}, and \textbf{Attributes} using this process. 

\paragraph{Metrics} 

We introduce variants of the ranking metric Hits@k, called Contradict@k and Entail@k.
\textbf{Contradict@k} measures the proportion of top-k candidates returned by the model which contradict candidates, averaged over examples. This measures the propensity of a model to highly rank contradictions. Contradiction@1 is the proportion of consistency errors made by the model. For this metric lower values are better, in contrast to Hits@k. 

\textbf{Entail@k} measures the proportion of top-k candidates returned by the model which are entailment candidates, averaged over examples. Entailment candidates share the same underlying triple as the ground-truth next utterance, so this metric rewards highly ranked candidates that convey similar meaning and logic to the ground-truth utterance. Thus it can be interpreted as a more permissive version of Hits@k.

\paragraph{Results}

Table~\ref{tbl:rerank} shows re-ranking results on the three evaluation sets ($\lambda=1.0, k=10$). The NLI re-ranking improves all three metrics on all the evaluation sets. Overall dialogue performance improves, as measured by Hits@1. The NLI re-ranking substantially reduces the number of contradicting utterances predicted by the model, and increases the number of utterances which entail a profile sentence, as seen in the Contradict@1 and Entail@1 scores.

Figure~\ref{fig:example} shows an example dialogue with candidates, contradictions predicted by the NLI model, and the corresponding re-ranked candidates.

\subsection{Experiment 3: Human Evaluation}

This experiment evaluates the effect of the proposed NLI re-ranking method on a dialogue model's consistency, where consistency is judged by human annotators in an interactive persona-based dialogue setting.

\paragraph{Experiment Setup} 

We use ParlAI~\cite{Miller2017ParlAI:Platform} which seamlessly integrates with Amazon Mechanical Turk for human evaluation. A human annotator is paired with a model, and each is randomly assigned a persona from 1,155 persona sets. The human and model are then asked to make a conversation of at least either five or six turns (randomly decided). After the conversation, the annotator assigns three scores to the conversation, described below. Each annotator is allowed to participate in at most ten conversations per model, and we collect 100 conversations per model. Two models are evaluated: the same key-value memory network used in Experiment~\ref{expr:dialogue} without re-ranking (\textbf{KV-Mem}), and with re-ranking (\textbf{KV-Mem + NLI}).

\paragraph{Scoring and Calibration} 

Following a conversation, an annotator is shown the conversation and the model's persona, and assigns three scores: an overall score of how well the model represented its persona (\{1,2,3,4,5\}), a marking of each model utterance that was consistent with the model's persona (\{0,1\}), and a marking of each model utterance that contradicted a previous utterance or the model's persona (\{0,1\}).

We Bayesian calibration to adjust for annotator bias, following \cite{Kulikov2018ImportanceModelling}. We assume a model with observed scores $S_{ij}$ and latent variables $M_i$ for the unobserved score of model $i$ and $B_j$ for the bias of annotator $j$. We then estimate the posterior mean and variance for the unobserved scores given the observed scores. See Appendix~\ref{apx:calibrate} for details.

\paragraph{Results} 

Table~\ref{tbl:human} shows the human evaluation results. The natural language inference re-ranking improves all the metrics, notably the fine-grained consistency score (0.27 vs. 0.35) and contradiction score (0.25 vs. 0.16). The results are consistent with the conclusions from the automatic evaluation in Experiment~\ref{expr:dialogue}. 

\section{Conclusion}

In this paper, we demonstrated that natural language inference can be used to improve performance on a downstream dialogue task. To do so, we created a new dialogue-derived dataset called Dialogue NLI, a re-ranking method for incorporating a Dialogue NLI model into a dialogue task, and an evaluation set which measures a model's persona consistency. The dataset offers a new domain for NLI models, and suggests avenues such as 
devising alternative methods for using natural language inference components in downstream tasks. 

\bibliographystyle{plain}
\bibliography{main.bib}

\appendix

\appendix
\newpage

\section{Dataset Details}
\subsection{Schema}
\paragraph{Relation Types}: \\
place\_origin,
live\_in\_citystatecountry,
live\_in\_general,
nationality,
employed\_by\_company,
employed\_by\_general,
has\_profession,
previous\_profession,
job\_status,
teach,
school\_status,
has\_degree,
attend\_school,
like\_general,
like\_food,
like\_drink,
like\_animal,
like\_movie,
like\_music,
like\_read,
like\_sports,
like\_watching,
like\_activity,
like\_goto,
dislike,
has\_hobby,
has\_ability,
member\_of,
want\_do,
want\_job,
want,
favorite\_food,
favorite\_color,
favorite\_book,
favorite\_movie,
favorite\_music,
favorite\_music\_artist,
favorite\_activity,
favorite\_drink,
favorite\_show,
favorite\_place,
favorite\_hobby,
favorite\_season,
favorite\_animal,
favorite\_sport,
favorite,
own,
have,
have\_pet,
have\_sibling,
have\_children,
have\_family,
have\_vehicle,
physical\_attribute,
misc\_attribute,
has\_age,
marital\_status,
gender,
other.

Additional triples with a not\_have relation were extracted using a dependency tree pattern.
\label{apx:schema}

\paragraph{Entity Categories}: ability,
activity,
animal,
color,
citystate,
country,
company,
cuisine,
degree\_type,
drink,
family,
food,
gender,
general\_location,
job\_status,
language,
marital,
media\_genres,
media\_other,
movie\_title,
music\_artist,
music\_genre,
music\_instrument,
noun,
number,
organization,
person,
person\_attribute,
person\_label,
personality\_trait,
profession,
read\_author,
read\_genre,
read\_title,
read\_other,
school\_name,
school\_status,
school\_type,
season,
sport\_type,
subject,
time,
vehicle,
location,
other.
\subsection{Relation Swaps}
Relation swaps for contradictions include \texttt{(have\_*, not\_have)}, \\\texttt{(own, not\_have)}, \\\texttt{(has\_hobby, not\_have)}, \\\texttt{(like\_*, dislike)}, \\\texttt{(favorite\_*, dislike)}.

Neutral relation swaps include \texttt{(have\_x, have\_y)}, e.g. \texttt{have\_pet}, \texttt{have\_sibling}. Additional \texttt{(have\_* A, not\_have B)} swaps were defined for entities A which are a super-type of B, namely (A,B) pairs (\{pet, animal\}, \{dog, cat\}), (\{sibling\}, \{brother, sister\}), (\{child, kid\}, \{son, daughter\}), (\{vehicle\}, \{car, truck\}); this includes sentence pairs such as ``i have a sibling'', ``i do not have a sister''. Similarly, \texttt{(not\_have B, have\_* A)} swaps were defined using the (A, B) pairs above.
\label{apx:relswap}
\subsection{Entity Swaps}

For contradictions, swapping entities for the following relation types was assumed to yield a contradiction:

attend\_school,
employed\_by\_company,
employed\_by\_general,
favorite\_animal,
favorite\_book,
favorite\_color,
favorite\_drink,
favorite\_food,
favorite\_hobby,
favorite\_movie,
favorite\_music,
favorite\_music\_artist,
favorite\_place,
favorite\_season,
favorite\_show,
favorite\_sport,
gender,
has\_profession,
job\_status,
live\_in\_citystatecountry,
marital\_status,
nationality,
place\_origin,
previous\_profession,
school\_status,
want\_job.

Additionally, for \texttt{physical\_attribute}, \texttt{misc\_attribute}, or \texttt{other} relations, an entity swap was done using all WordNet antonym pairs in the personality\_trait and person\_attribute entity categories, as well as the swaps (\{blonde\}, \{brunette\}), (\{large\}, \{tiny\}), (\{carnivore, omnivore\}, \{vegan, vegetarian\}), (\{depressed\}, \{happy, cheerful\}), (\{clean\}, \{dirty\}) where each entity in the left set is swapped with each entity in the right set.

\label{apx:eswap}

\section{Experiment Details}
\paragraph{Experiment 1} The InferSent model used the Adam \cite{Kingma2014Adam:Optimization} optimizer with learning rate 0.001, and otherwise used the hyper-parameters from the open source implementation\footnote{https://github.com/facebookresearch/InferSent}. The ESIM model used a 1-layer bidirectional LSTM with hidden dimension 1024 and Adam optimizer with learning rate 0.0001, with the remaining hyper-parameters set to those used by the InferSent model.

\paragraph{Experiment 2} The dialogue model was trained using ParlAI \cite{Miller2017ParlAI:Platform} on the \texttt{personachat:self\_original} task, using the  hyper-parameters given for the \texttt{KVMemnnAgent} in the ConvAI2 competition. The NLI model was the same ESIM model from Experiment 1.

\section{Score Calibration}
\label{apx:calibrate}

\paragraph{1-5 star rating}

Let $M_i \sim \mathcal{N}(\mu_i, 1^2)$ be the unobserved, underlying quality of the $i$-th approach, where $\mu_i \sim \mathcal{U}(1,5)$. Let $A_j \sim \mathcal{N}(0, 1^2)$ be the unobserved annotator bias, indicating whether the $j$-th annotator is more or less generous. We observe a score given by the $j$-th annotator to the $i$-th approach, and this score follows a normal distribution with its mean given by the sum of the underlying model score and annoator bias, i.e., $S_{ij} \sim \mathcal{N}(M_i + A_j, 1^2)$. We observe some of these scores, and given these scores, the goal is to infer $\mathbb{E}[M_i]$ and $\mathbb{V}[M_i]$ for all $i$.

\paragraph{Utterance-pair selection}

Each annotator is asked to label each utterance-pair as consistent and/or contradictory with respect to the personas. In this case, the unobserved, underlying model score is modelled as a pre-sigmoid normal variable, i.e., $M_i \sim \mathcal{N}(0, 1^2)$, and the annotator bias as a usual normal variable, i.e., $A_j \sim \mathcal{N}(0, 1^2)$, similarly to the 1-5 star rating case above. We however also introduce a turn bias $T_k \sim \mathcal{N}(0, 1^2)$ to incorporate the potential degradation of a neural dialogue model as the conversation lengthens. An observed score for each utterance pair then follows a Bernoulli distribution with its mean given as the sigmoid of the sum of these three latent variables, i.e., $S_{ijk} \sim \mathcal{B}(\text{sigmoid}(M_i+A_j+T_k))$. The goal of inference is to compute $\mathbb{E}[\text{sigmoid}(M_i)]$ and $\mathbb{V}[\text{sigmoid}(M_i)]$.

\paragraph{Inference}

We use Pyro\cite{Bingham2018Pyro:Programming} and the no-u-turn sampler (NUTS)\cite{Hoffman2014TheCarlo} for posterior inference.

\end{document}